\title{KBP}
\documentclass[11pt,letterpaper]{article}
\usepackage{ijcnlp2017}
\usepackage{times}
\usepackage{color}
\usepackage{latexsym}
\usepackage{graphicx}
\usepackage{tikz}
\def\checkmark{\tikz\fill[scale=0.2](0,.35) -- (.25,0) -- (1,.7) -- (.25,.15) -- cycle;} 
\ijcnlpfinalcopy

\def\ijcnlppaperid{***}

\title{TAMU at KBP 2017:  Event Nugget Detection and Coreference Resolution}

 \author{Prafulla Kumar Choubey \and Ruihong Huang \\
         Department of Computer Science and Engineering\\
		Texas A\&M University\\
         {\tt (prafulla.choubey, huangrh)@tamu.edu}}

\date{}

\begin{document}

\maketitle

\begin{abstract}
In this paper, we describe TAMU's system submitted to the TAC KBP 2017 event nugget detection and coreference resolution task. Our system builds on the statistical and empirical observations made on training and development data. We found that modifiers of event nuggets tend to have unique syntactic distribution. Their parts-of-speech tags and dependency relations provides them essential characteristics that are useful in identifying their span and also defining their types and realis status. We further found that the joint modeling of event span detection and realis status identification performs better than the individual models for both tasks. Our simple system designed using minimal features achieved the micro-average F1 scores of 57.72, 44.27 and 42.47 for event span detection, type identification and realis status classification tasks respectively. Also, our system achieved the CoNLL F1 score of 27.20 in event coreference resolution task. 

\end{abstract}

\section{Introduction}
The TAMU NLP group participated in the Event Nugget Track of TAC KBP 2017. The goal of this track is to identify the character span, classify type and realis status of event mentions and also link all the coreferent event mentions within the same text. We designed a pipeline of three neural network based classifiers for this task, the first detects event span and classify realis status, the second classifies event type and the third resolves event coreference links. These classifiers are based on simple lexical and syntactic features which are derived from the distinct distributional properties of event mentions.

Syntactic dependency relation of event triggers with their modifiers and governors are lately shown very effective for the task of temporal relations classification between event pairs ~\cite{choubey2017sequential, yao2017weakly, cheng-miyao:2017:Short} and identifying the temporal status of an event mention ~\cite{dai2017using}. The realis status of an event mention has a close association to its temporal status \cite{huang-EtAl:2016:EMNLP2016} and its relative position in temporal space. Motivated by the performance gain observed in recent research works on temporal relations, we analyze the distribution of modifiers of events with different realis status. 
Let's look at the examples below ({\fontfamily{cmr}\selectfont boldfaced words in blue are event mentions and other words in blue are their modifiers}):

(1) {\bf [Actual]} {\it Continental Airlines {\color{blue} board} of directors {\color{blue} \bf met} {\color{blue}Wednesday} to {\color{blue}discuss} a  merger with United Airlines, a person familiar with the situation said.}

(2) {\bf [Other]} { {\it \color{blue} If United} and Continental {\color{blue} \bf marry}, the new airline will be the nation's largest carrier, eclipsing Delta Airlines, which merged with Northwest Airlines in 2008.}

(3) {\bf [Actual]} { If United and Continental  marry, the new airline will be the nation's largest carrier, eclipsing Delta {\color{blue}Airlines}, which {\color{blue} \bf merged} with Northwest {\color{blue}Airlines} in {\color{blue}2008}.}

In examples (1) and (3), the presence of modifiers {\bf Wednesday} and {\bf 2008} help in binding the events {\it met} and {\it merged} to the timeline. These temporal modifiers imply that both the events have already occurred in the past and thus should be classified as the {\bf actual} event. On the other hand, the modifier {\bf if} of event {\it marry} in example (2) implies that the event is hypothetical. Our analysis and empirical evaluation suggest that these dependency parse based features are also beneficial to identifying the realis status of events. 

In our experiments, we further found that the event span detection performs better when modeled jointly with realis status identification. We evaluated two neural network classifiers- the first classifier is trained to predict whether the given word is an event trigger or not and the second classifier is trained to jointly predict whether the given word is an event trigger together with its realis status on the {\it 2016 evaluation dataset}. We found that the second classifier achieved around 2\% higher F1 score on event span detection, with major improvement coming from the precision.

\section{Motivation}

\begin{table}[h]
\small
\begin{center}
\begin{tabular}{|l|l|l|l|l|}
\hline \bf Dep. Rel. & \bf Actual  & \bf Generic  & \bf Other & \bf Non-Event \\ \hline
nsubjpass & \bf 307 & 51 & 153 & 729\\
ccomp & \bf 305 & 30 & 54 & 2266\\
nmod:in & \bf 316 & 29 & 67 & 1456\\
mark & 327 & 117 & \bf 418 & 3876\\
auxpass & \bf 336 & 67 & 180 & 1422\\
dobj & \bf 671 & 154 & 495 & 5329\\
nmod:tmod & \bf 114 & 4 & 18 & 302\\
nmod:into & \bf 23 & 3 & 11 & 66\\
nmod:agent & \bf 37 & 5 & 16 & 115\\
compound & \bf 260 & 87 & 85 & 5694\\
dep & \bf 106 & 47 & 53 & 3078\\
\hline
\end{tabular}
\end{center}
\caption{\label{deprel} Frequency of dependency relation with modifiers for event and non-event words }
\end{table}

We analyzed the dependency parse of sentences and found that modifiers of event trigger word have unique syntactic distribution. They are related to the trigger word with few frequently occurring dependency relations. Moreover, they tend to have few specific parts-of-speech (POS) tags only. Based on our observations on {\it 2015 training data}, words having a modifier with a set of dependency relations like {\it ccomp, nmod:in, nmod:tmod, nsubjpass, auxpass etc.} are event triggers with very high probability. At the same time, words having modifiers attached with other relations like {\it compound, dep, etc.} are almost always non-event words (Table~\ref{deprel}). Similar distribution also holds with the parts-of-speech tags of modifiers. While some of the POS-tags including {\it WP, VBD, IN, TO etc.} are frequently associated with event triggers, other POS-tags like {\it EX, POS etc.} are common to the non-event words (Table~\ref{pos}).

\begin{table}[h]
\small
\begin{center}
\begin{tabular}{|l|l|l|l|l|}
\hline \bf POS & \bf Actual  & \bf Generic  & \bf Other & \bf Non-Event \\ \hline
WP & \bf 99 & 23 & 3 & 655\\
RP & \bf 49 & 15 & 39 & 423\\
MD & 36 & 45 & \bf 254 & 1427\\
NNP & \bf 1108 & 52 & 233 & 7538\\
VBD & \bf 594 & 28 & 120 & 2592\\
PRP & \bf 494 & 108 & 419 & 5987\\
TO & 109 & 85 & \bf 282 & 2803\\
IN & \bf 809 & 283 & 354 & 12271\\
EX & 3 & \bf 4 & \bf 4 & 224\\
POS & \bf 5 & 0 & 3 & 680\\
\hline
\end{tabular}
\end{center}
\caption{\label{pos} Frequency of parts-of-speech tags of modifiers for event and non-event words}
\end{table}

We further analyzed the distribution of POS-tags of words in the surface context of event words(Table~\ref{pos-cont}). On comparing the ratio of frequencies of various POS-tags w.r.t. event and non-event words in Table~\ref{pos} and \ref{pos-cont}, it is evident that context defined on words along dependency path is more informative than the neighbor words along the surface path.

\begin{table}[h]
\small
\begin{center}
\begin{tabular}{|l|l|l|l|l|}
\hline \bf POS & \bf Actual  & \bf Generic  & \bf Other & \bf Non-Event \\ \hline
WP & \bf 90 & 17 & 4 & 3088\\
RP & \bf 77 & 34 & 58 & 1870\\
MD & 33 & 45 & \bf 239 & 6728\\
NNP & \bf 1027 & 107 & 147 & 33982\\
VBD & \bf 1405 & 73 & 189 & 16978\\
TO & \bf 326 & 122 & 40 & 12228\\
POS & \bf 92 & 11 & 12 & 2630\\
\hline
\end{tabular}
\end{center}
\caption{\label{pos-cont} Frequency of parts-of-speech tags of words in surface context of event and non-event words}
\end{table}

We also analyzed the distribution of name entities that modify event triggers in the syntactic parse tree. Since each type of event participants can only be linked to specific event subtypes only, named entities are a strong feature for type classification. The distribution is shown in Table~\ref{ner}. 
Clearly, each type of event tends to feature certain types of entities as arguments, therefore, 
the presence of entities can serve as a useful evidence for event type classification.

\begin{table}[h]
\small
\begin{center}
\begin{tabular}{|l|l|l|l|l|l|}
\hline \bf Event Type & \bf Per.  & \bf Loc.  & \bf Org. & \bf Num. & \bf Misc. \\ \hline
Elect & 12 & 1 & 4 & 0 & 0\\
Pardon & \bf 41 & 0 & 0 & 5 & 0\\
Sentence & \bf 16 & 3 & 1 & 0 & 0\\
Start-position & \bf 18 & 1 & 2 & 0& 0\\
End-Org. & 0 & 0 & 3 & 0 & 0\\
Transfer-money &  \bf 10 & 4 & \bf 14 & 1 & 0\\
Transport-art. & 0 & \bf 13 & 1 & 0 & 0\\
Attack & \bf 14 & \bf 60 & 3 & 5 & 8 \\
Broadcast & \bf 39 & \bf 11 & \bf 28 & 1 & 0\\
Demonstrate & 0 & \bf 13 & 2 & 2 & 1\\
Transport-person &  \bf31 & \bf 66 & 1 & 8 &1 \\
Contact & \bf 53 & 7 & \bf 12 & 1 & 1 \\
Die & \bf 37 & \bf 11 & 4 & 4 & 3\\
Meet & \bf 36 & 4 & 6 & 0 & 0\\
Acquit & 1 & 0 & 0 & 0 & 0\\
\hline
\end{tabular}
\end{center}
\caption{\label{ner} Distribution of named entities types in the context of various event subtypes ({\fontfamily{cmr}\selectfont Per.- person, Loc.- location, Org.- organization, Num.-number and Misc.- miscellaneous})}
\end{table}

\section{System Overview}
Our feature based method follows the conventional pipeline approaches which divide event nugget detection and coreference resolution into several sub-tasks\footnote{Implementation is available at \url{https://github.com/prafulla77/TAC-KBP-2017-Participation}}. These are:

\begin{table}[h]
\small
\begin{center}
\begin{tabular}{|l|l|l|l|}
\hline \bf Features & \bf Dim. & \bf S+R  & \bf T \\ \hline
lemma vector & 300 & \checkmark & \checkmark \\
token vector & 300 &  & \checkmark \\
POS-tag & 47 & \checkmark  &\\
context words POS-tag & 235 & \checkmark  &\\
context words dependency relation & 1040 & \checkmark  &\\
(token - lemma) vector & 300 & \checkmark  &\\
dependency relation with modifiers & 208 & \checkmark  & \checkmark  \\
POS-tag of modifiers & 47 & \checkmark  &\\
dependency relation with governor & 208 & \checkmark  &\\
POS-tag of governor & 47 & \checkmark  &\\
prefix and suffix of words & 36 & \checkmark  & \checkmark  \\
named entity type of modifiers & 8 & & \checkmark \\
\hline
\end{tabular}
\end{center}
\caption{\label{features} Features and their vector dimensions used in our span+realis and subtype classifiers ({\fontfamily{cmr}\selectfont S+R- span+realis, T- type}, Context features are defined over window of size 2)}
\end{table}

\subsection{Span identification and Realis Status Classification} In the first step, we jointly perform span identification and realis status classification. We use an ensemble of neural network classifiers defined over features described in Table~\ref{features}. All neural classifiers perform classification over 4 classes- actual event, generic event, other event and non-event. However, they differ to each other in terms of various hyper-parameters including the number of layers, number of neurons in each layer and dropout and activation function for each layer. This is done to reduce the variance and obtain more consistent results across datasets. The output layer in all neural network classifiers use softmax activation function and thus predict the probabilistic score for each class. The output scores from all the classifiers are directly added to obtain the final probability for each class and the aggregated probability is used to make the final decision.

\subsection{Event Subtype Classification}
Following the strategy similar to span detection and realis classification, event subtype classifier also uses an ensemble of classifiers defined over features described in Table~\ref{features}. We used KBP 2015 training and evaluation dataset to train our system. However, that dataset contains 38 event subtypes while the KBP 2017 evaluation dataset contains events from 18 subtypes only. So we model this subtask as a 19 class classification problem, where 19 classes correspond to the 18 subtypes in KBP 2017 evaluation dataset and the {\it other}. The other class means that event can be from any of the remaining 20 subtypes that are not included in evaluation dataset. Also, there are several event mentions in the dataset that have multiple subtypes. We consider only one subtype for such event mentions and ignore other subtype instances.

We trained 10 neural network classifiers for span detection and realis status identification and 3 classifiers for type classification. These classifiers differ in their architecture, training parameters and initialization. Details of all the classifiers used are described in Table~\ref{clf}. The configuration  [{\it 2468-600-600-50-4, 0-.5-0-0-0, 10}] can be interpreted as a classifier with an input layer with 2468 neurons, 3 hidden layers with 600, 600 and 50 neurons and an output layer with 4 neurons. The classifier has a dropout layer (with the dropout rate of 0.5) after first hidden layer and is trained for 10 epochs. All the classifiers use {\it relu} activation in input layer, {\it tanh} activation in all hidden layers and {\it softmax} activation in output layers.

\begin{table}[h]
\small
\begin{center}
\resizebox{0.47\textwidth}{!}{\begin{tabular}{|l|l|l|l|}
\hline \bf Name & \bf Layers  & \bf Dropout  & \bf Epochs  \\ \hline
S 1 & 2468-600-600-50-4 & 0-.5-0-0-0 &  10  \\
S 2&  2468-600-600-50-4 & 0-.2-0-0 &  15  \\
S 3&  2468-2468-1234-600-200-4 & 0-.2-.5-.2-0 &  10  \\
S 4 & 2468-2468-1234-600-200-4 & 0-.2-.5-.2-0 &  15  \\
S 5&  2468-2468-1234-600-200-4 & 0-0-.5-.2-0 &  10  \\
S 6&  2468-2468-1234-600-200-4 & 0-0-.5-.5-0 &  15  \\
S 7 & 2468-2468-1234-600-200-4 & 0-0-.2-.2-0 &  15  \\
S 8&  2468-2468-1234-600-200-4 & 0-.5-.5-.5-0 &  15  \\
S 9&  2468-1000-600-200-4 & 0-.5-0-0 &  10  \\
S 10&  2468-1000-600-200-4 & 0-.5-0-0 &  15  \\
T 1 & 852-852-852-200-19 & 0-0-0-0 &  10  \\
T 2&  852-852-852-200-19 & 0-0-0-0 &  15  \\
T 3&  852-852-400-200-19 & 0-0-0-0 &  15  \\
\hline
\end{tabular}}
\end{center}
\caption{\label{clf} Span detection and realis status classifiers and type classifiers parameters. S means span+realis classifier and T means type classifier }
\end{table}

\subsection{Coreference Resolution} We replicated the pairwise within-document classifier architecture proposed in \citet{choubey2017event} for this task. The classifier uses a common neural layer shared between two event mentions that embed event lemma and parts-of-speech tags and then calculates cosine similarity, absolute and Euclidean distances between two event embeddings, corresponding to each event mention. This shared layer has 347 neurons and uses {\it sigmoid} activation function. The classifier also includes a second neural layer with 380 neurons to embed event arguments (considered named entities which modifies event mentions as the argument~\cite{finkel2005incorporating}) that are overlapped between the two event mentions, suffix and prefix based features for both event lemmas and absolute difference between vectors of event tokens. The calculated embeddings similarities as well as the embedding of the second neural layer are concatenated and fed into the third neural layer with 10 neurons. The output activation of the third layer is finally fed into the output layer with 1 neuron that gives the confidence score to indicate the similarity between the two event mentions\footnote{We implemented our classifier using the Keras library~\cite{chollet2015keras}}. The second, third and output layers also use {\it sigmoid} activation function. We used 300 dimensional word embeddings \cite{pennington2014glove} and 47 dimensional one hot embeddings for pos-tags \cite{toutanova03}. During inference, we perform greedy merging using the classifier’s predicted score. An event mention is merged to its best matching antecedent event mention if the predicted score is greater than 0.5.

\section{Experiments}

\subsection{Dataset}
The testing data of KBP 2017 consists of documents taken from the discussion forum and news articles. Therefore, we train our classifiers on both discussion forum and news articles taken from KBP 2015 training and evaluation dataset and used documents from KBP 2016 evaluation as the development dataset.

\subsection{Preprocessing}
We run Stanford coreNLP module for tokenization, sentence segmentation, lemmatization, POS tagging, dependency parsing, named entity recognition and coreference resolution \cite{manning-EtAl:2014:P14-5,recasens_demarneffe_potts2013, lee11conllst}. Further, we use the cleanxml annotator available in coreNLP pipeline for removing tags and obtaining character offsets of each token. The obtained offsets are aligned to the character offset provided in the annotation files.

\subsection{ Performance comparison on the development dataset}
In order to compare our system with the systems that participated in the event nugget detection and coreference task in  KBP 2016, we evaluated our system on KBP 2016 testing dataset and used it for development and parameter tuning. 

In Table~\ref{indiveval2016}, we illustrate the advantage of jointly modeling event span detection and realis status identification over their individual models. The results mentioned in the Table~\ref{indiveval2016} are the average F1 score of 3 classifiers' instances trained with different random initializations. From the table, it's evident that the average performance on span detection has improved significantly when modeled together with the realis status classification. However, the performance on realis status classification remains similar.

\begin{table}[h]
\small
\begin{center}
\resizebox{0.47\textwidth}{!}{\begin{tabular}{|l|l|l|}
\hline \bf System & \bf Span  & \bf Realis   \\ \hline
Joint span + realis classifier & \bf 53.47 & \bf 40.13  \\
Separate realis and span classifiers & 51.44 & 39.87 \\
\hline
\end{tabular}}
\end{center}
\caption{\label{indiveval2016} Performance comparison of joint span+realis and separate span and realis classifiers on KBP 2016 evaluation dataset}
\end{table}

\begin{table}[h]
\small
\begin{center}
\resizebox{0.47\textwidth}{!}{\begin{tabular}{|l|l|l|l|l|}
\hline \bf System & \bf Span  & \bf Type  & \bf Realis & \bf All \\ \hline
Ensemble System & \bf 56.14 & \bf 44.48 & \bf 42.59 & \bf 33.0 \\
Weakest Classifiers& 52.44 & 41.82 & 37.16 & 29.18 \\
Strongest Classifiers& 54.03 & 43.60 & 40.28 & 31.92\\
\hline
\end{tabular}}
\end{center}
\caption{\label{ensemble2016} Performance comparison of our ensemble system with the best and the worst member classifiers on KBP 2016 evaluation dataset}
\end{table}

In Table~\ref{ensemble2016}, we compare the performance of our ensemble based model with its strongest and weakest member classifiers. The results show that combining multiple classifiers helped overcome the inherent problem of the neural network to over-fit according to the specific dataset. Including diverse classifiers with different dropout and network architecture helped reduce variance in the final prediction. 

In Table~\ref{eval2016} and \ref{evalcoref2016}, we compare the performance of our complete model with the systems submitted to the KBP 2016. Our feature based classifier compares well to the top scoring systems in KBP 2016 which modeled this task as sequence labeling problem and used complex models based on recurrent neural networks and convolutional neural networks. Specifically, compared to the best scores in KBP 2016, our model is able to achieve around 1.5\% higher F1 score in event span detection task and is marginally below the best score in realis status classification task. This implies the advantage of using the dependency parse based features and joint modeling of event span detection and realis status classification subtasks.

\begin{table}[h]
\small
\begin{center}
\resizebox{0.47\textwidth}{!}{\begin{tabular}{|l|l|l|l|l|}
\hline \bf System & \bf Span  & \bf Type  & \bf Realis & \bf All \\ \hline
Our System & \bf 56.14 &  44.48 & 42.59 & 33.0 \\
\citet{lu2016utd}& 54.59 & \bf 46.99 & 39.78 & 33.58 \\
\citet{nguyen2016new}& 54.07 & 44.38 & \bf 42.68 & \bf 35.24\\
\citet{hong2016kbp}& 50.83 & 43.67 & 38.35 & 32.59 \\
\citet{liu2016cmu}& 50.49 & 44.61 & 33.11 & 29.06 \\
\citet{wipsys2016}& 49.39 & 44.47 & 36.96 & 33.1 \\
\citet{yurpi}& 48.65 & 42.07 & 34.46 & 30.16\\
\citet{TUD-CS-2016-14752}& 46.85 & 32.62 & 36.83 & 26.53\\
\citet{weioverview}& 43.33 & 36.70& 33.69 & 28.38\\
\citet{washi2016KBP}& 41.25 & 34.65 & 29.75 & 25.24\\
\citet{satyapanich2016event}& 35.24 & 31.57 & 24.04 & 21.67\\
\citet{yang2016micro}& 29.21 & 24.77 & 21.13 & 17.87\\
\citet{TMPSMRR16}& 28.07 & 21.57 & 9.70 & 7.49\\
\citet{disc}& 5.72 & 0.59 & 2.75 & 0.11\\
\hline
\end{tabular}}
\end{center}
\caption{\label{eval2016} Performance comparison of our system on event span, type and realis status classification w.r.t. systems submitted in KBP 2016. All results are taken from ~\citet{mitamura2016overview}}
\end{table}

\begin{table}[h]
\small
\begin{center}
\resizebox{0.47\textwidth}{!}{\begin{tabular}{|l|l|l|l|l|l|}
\hline \bf System & \bf $B_3$  & \bf CeafE  & \bf MUC & \bf BLANC & \bf CoNLL \\ \hline
Our System & 36.62 &  \bf 35.50 & 17.62 & 18.77 & 27.13 \\
\citet{lu2016utd}& \bf 37.49 & 34.21 & \bf 26.37 & \bf 22.25 & \bf 30.08 \\
\citet{liu2016cmu} & 35.06 & 30.45 & 24.60 & 18.79 & 27.23\\
\citet{nguyen2016new} & 34.62 & 33.33 & 22.01 & 18.31 & 27.07\\
\citet{yurpi} & 20.96 & 16.14 & 17.32 & 10.67 & 16.27 \\
\citet{yang2016micro} & 19.74 & 16.13 & 16.05 & 8.92 & 15.21\\
\citet{TMPSMRR16} & 11.92 & 11.54 & 4.34 & 3.10 & 7.73\\
\hline
\end{tabular}}
\end{center}
\caption{\label{evalcoref2016}  Performance comparison of our system on coreference resolution w.r.t. systems submitted in KBP 2016. All results are taken from ~\citet{mitamura2016overview}}
\end{table}

\section{Evaluation on KBP 2017 dataset}
We submitted 3 runs of our system for the official evaluation. They are:

\noindent \underline{Run-I}: used ensemble of classifiers without any parameter tuning.

\noindent \underline{Run-II}: same as Run-I with parameters tuned to produce the best result on 2016 Evaluation dataset.

\noindent \underline{Run-III}: used the strongest member classifier from all the classifiers used for event span, realis status and type classification in Run-I. The coreference resolution classifier is same in all three runs.

\begin{table*}[h]
\small
\begin{center}
\resizebox{0.97\textwidth}{!}{\begin{tabular}{|l|l|l|l|l|l|l|l|l|l|l|l|l|}
\hline \bf Runs & \bf Span-P  & \bf Span-R  & \bf Span-F1  & \bf Type-P  & \bf Type-R  & \bf Type-F1  & \bf Realis-P  & \bf Realis-R  & \bf Realis-F1  & \bf All-P  & \bf All-R  & \bf All-F1  \\ \hline
I & 58.95 & \bf 56.53 & \bf 57.72 & 45.21 & 43.36 & 44.27 & 43.38 & \bf 41.60 & \bf 42.47 & 32.64 & 31.31 & 31.96\\
II& \bf 64.22 & 50.45 & 56.50 & \bf 50.32 & 39.53 & 44.28 & \bf 47.30 & 37.16 & 41.62 & \bf 36.30 & 28.52 & 31.94 \\
III& 57.44 & 54.44 & 55.90 & 45.88 & \bf 43.48 & \bf 44.65 & 42.07 & 39.87 & 40.94 & 33.35 & \bf 31.60 & \bf 32.45 \\
\hline
\end{tabular}}
\end{center}
\caption{\label{eval2017}  Performance of our system for span, type and realis classification on KBP 2017 evaluation dataset [results released by the organizers]}
\end{table*}

\begin{table}[h]
\small
\begin{center}
\resizebox{0.47\textwidth}{!}{\begin{tabular}{|l|l|l|l|l|l|}
\hline \bf Runs & \bf $B_3$  & \bf CeafE  & \bf MUC & \bf BLANC & \bf CoNLL \\ \hline
I & 35.0 &  \bf \bf 34.95 & 18.66 & 16.54 & 26.29 \\
II & 34.34 & 33.63 & \bf \bf 22.90 & \bf \bf 17.94 & \bf \bf 27.20 \\
III & \bf \bf 35.03 &  34.67 & 18.68 & 16.47 & 26.21 \\
\hline
\end{tabular}}
\end{center}
\caption{\label{evalcoref2017}  Performance comparison of our system on coreference resolution [results released by the organizers]}
\end{table}

Comparison of results of 3 runs (Tables \ref{eval2017} and \ref{evalcoref2017}) shows mixed performance. However, our hypotheses are consistent. They key observations are:
\begin{enumerate}
\item Run I achieves the highest F1 score for span detection, realis status classification and realis + type classification. This model doesn't use any form of tuning on the development dataset. We can arguably conclude that inference made by aggregating multiple diverse classifiers can reduce dependency on the training parameters like dropout rates, layers etc. in the Neural Networks.
\item The events extracted in run II achieved the best coreference performance. This can be justified by the coreference evaluation setup, which requires the coreferent event mention to have the same event type. Run II has significantly higher precision for all the subtasks- span, type and realis. 
\item Similar to the results on development dataset, ensemble based system (run I) performs better than the system relying on single classifier for each subtask (run III).
\end{enumerate}

\subsection{Macro analysis of Results}
\begin{figure}[t]
  \centering \includegraphics [width=0.471\textwidth,keepaspectratio]{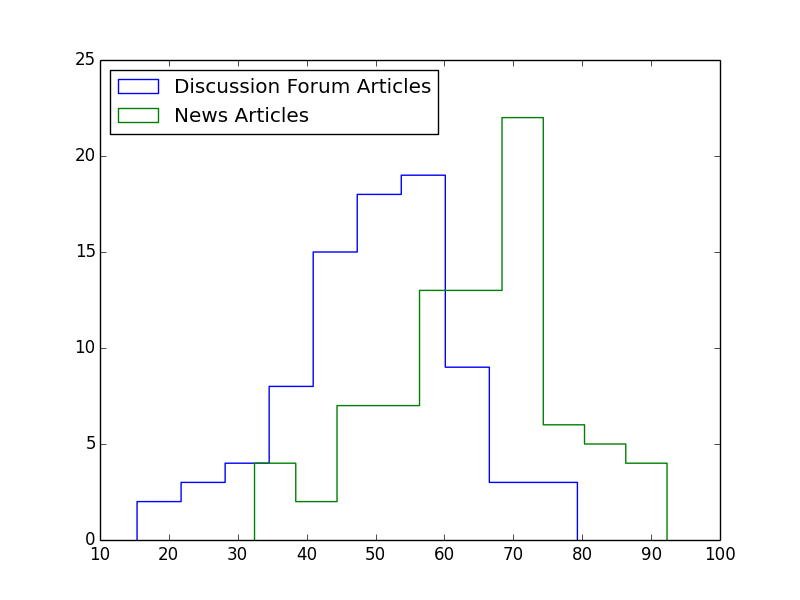}
  \caption{Number of Documents vs F1 score for Event Span Detection subtask}
  \label{span}
\end{figure}

\begin{figure}[t]
  \centering \includegraphics [width=0.471\textwidth,keepaspectratio]{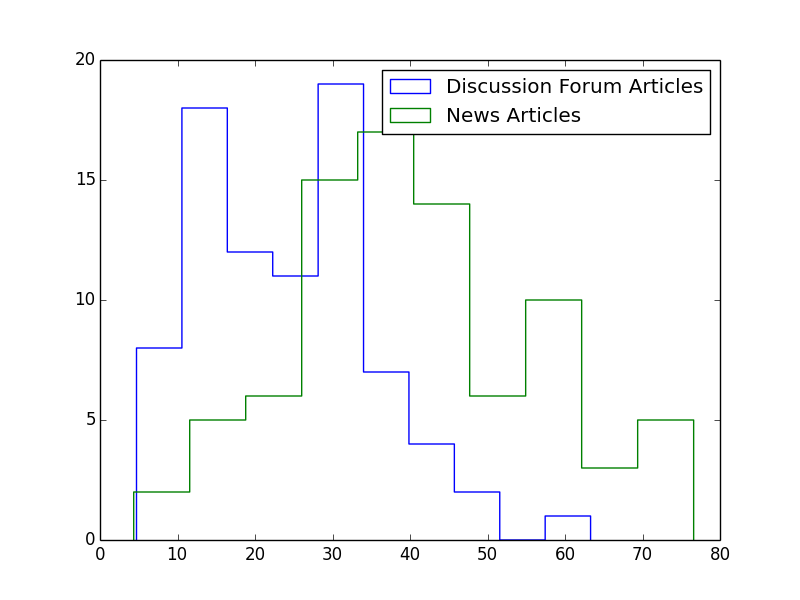}
  \caption{Number of Documents vs F1 score for Event Type and Realis Status subtask}
  \label{all}
\end{figure}

The KBP 2017 evaluation dataset contains two types of documents- discussion forum and news articles. While news articles are well structured, discussion forum articles are informal and noisy. Discussion forum articles tend to contain unnecessary punctuations, or sometimes omit punctuations and have several grammatical and spelling mistakes. Since our classifiers rely heavily on the features derived from syntactic parse, we separately analyzed the performance of our system on the discussion forum and news articles. Figures \ref{span} and \ref{all} shows the histogram of documents vs F1 score for span detection and type + realis status classification subtasks. It is quite interesting to find here that our system performed significantly better for the news articles compared to the noisy discussion forum articles. The lower performance of our systems on discussion forum articles can be partially accounted to the error generated from the preprocessing step. We manually analyzed the output from our preprocessing steps and observed that incorrect sentence segmentation is the dominant source of errors in most of the documents. The incorrect sentence segmentation abruptly changes the dependency parse tree that lowers our system's performance.

\subsection{Conclusion and Future work}
In this paper, we described TAMU's participation in TAC KBP 2017 event nugget and coreference track. Our feature based system showed the advantage of using dependency parse tree based features for this task. Empirically, we also found that the joint modeling of event span detection and realis status identification helps in improving the performance. This is particularly interesting and we plan to continue our work in this direction.

\bibliography{ijcnlp2017}
\bibliographystyle{ijcnlp2017}

\end{document}